\documentclass[journal]{IEEEtran}

\usepackage{graphics} 
\usepackage{epsfig} 
\usepackage{mathptmx} 
\usepackage{times} 
\usepackage{amsmath} 
\usepackage{amssymb}  

\ifCLASSINFOpdf
\else
\fi
\begin{document}
%
\title{Minimalist Regression Network with Reinforced Gradients and Weighted Estimates: a Case Study on Parameters Estimation in Automated Welding}
%
%
%

\author{Soheil Keshmiri
\thanks{Soheil Keshmiri is with Hiroshi Ishiguro Laboratories (HIL), Advanced Telecommunications Research Institute International (ATR), Kyoto, Japan, soheil@atr.jp}}
\maketitle

\begin{abstract}
This paper presents a minimalist neural regression network as an aggregate of independent identical regression blocks that are trained simultaneously. Moreover, it introduces a new multiplicative parameter, shared by all the neural units of a given layer, to maintain the quality of its gradients. Furthermore, it increases its estimation accuracy via learning a weight factor whose quantity captures the redundancy between the estimated and actual values at each training iteration. We choose the estimation of the direct weld parameters of different welding techniques to show a significant improvement in calculation
of these parameters by our model in contrast to state-of-the-arts techniques in the literature. Furthermore, we demonstrate the ability of our model to retain its performance when presented with combined data of different welding techniques. This is a nontrivial result in attaining an scalable model whose quality of estimation is independent of adopted welding techniques.
\end{abstract}


%
\IEEEpeerreviewmaketitle

\section{Introduction}
\label{intro}
\subsection{Overview}
\label{overview}
\IEEEPARstart{A}{utomation} of the welding processes witness a significant advancement in recent decades. This, in turn,  introduces a dramatic  improvement on the manufacturing productivity and its efficiency. As the interest in automation of this field of industry continues to grow, the needs for quality research on reducing the manual modification of its parameters by human operator becomes indispensable. For instance, most welding tasks involve fabrication of a large number of joints over an extended period of time where many of these joints require multiple welding passes. Furthermore, these tasks are mostly manual processes where the completion of a multi-pass welding for a single joint takes approximately a work shift, on average, by highly skilled welders. This limits the productivity of companies since there exists a limited number of welders with such a skill. Moreover, many of such welding tasks take place in outdoor environment whose conditions are hostile to human health. Therefore, the automation of this fabrication process is highly desirable.\newline
\begin{figure}[b!]
\begin{center}
\includegraphics[height=1.2in,width=3.2in]{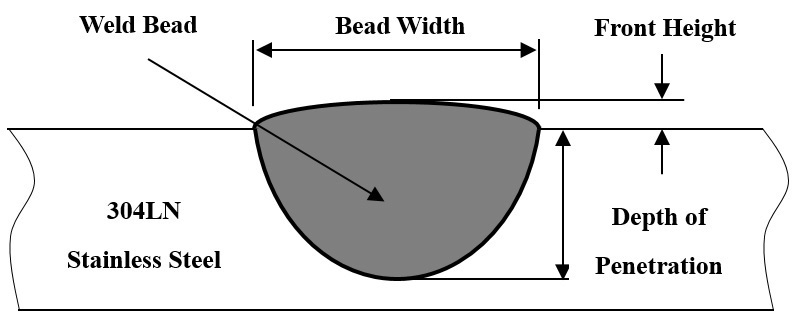}
\caption{Bead width and depth of penetration as examples of Direct Weld Parameters (DWP). The weld bead along with its front height on 304LN stainless steel is shown in the figure.} 
\label{DWP}
\end{center}
\end{figure} 
The intelligent monitoring of the welding process helps increase the precision and accuracy of the various aspects of its final outcome (e.g., the geometry of the weld beads). Furthermore, it helps determine if the driving parameters of the welding system requires readjustment. As a result, the overall performance of the system and consequently its welding quality improves significantly.\newline
Welding is an integral part of virtually every industrial process, from construction of buildings and complex structures to automotive and shipyard production lines. This explains the existence of a wide range of welding techniques from Gas Metal Arc Welding (GMAW)~\cite{kimsonyang2003} and Sub-merged Arc Welding (SAW)~\cite{karaoglusecgin2008} to shielded metal-arc welding (SMAW)~\cite{NageshDatta2002}, Tungsten Insert Gas Welding (TIG)~\cite{xudongwei2007}, and Gas Tungsten Arc Welding (GTAW)~\cite{andersonramaswamy1990}.\newline
A welding task, in its common form, comprises of a number of input control signals. Some examples include the voltage, the current, the torch traveling speed, the wire feed rate, and the arc gap. Cook~\cite{Cook2002} uses the term Indirect Weld Parameters (IWP) to refer to these input control signals. Furthermore, Nunes~\cite{nunes1983} shows that IWP directly influence the geometrical description of the welding pool i.e., its Direct Weld Parameters (DWP)~\cite{Cook2002}. The depth of penetration, the width of weld beads, and the transverse cross-sectional areas are some of the examples of these parameters. Firgure~\ref{DWP} illustrates these parameters for a weld bead on a 304LN stainless steel. A thorough investigation of the relationship between DWP and IWP is presented by Chandel and Bala~\cite{Chandel1998}.\newline
Research pertaining to welding processes is broadly categorized into two domain of studies, namely the estimation of the weld parameters and their sensitivity analysis. The latter refers to the study of the effect of the changes of IWP on the final weld quality. Whereas, the estimation problem attempts to predict the weld parameters, given the IWP. In this article, we focus on the estimation problem.\newline
Rosenthal~\cite{Rosenthal1941} presents an early static model based on heat flow where the welding contour is expressed as a function of electrode velocity, heat input, and the material properties. In addition, many researchers apply sophisticated mathematical models to capture this relationship. Some examples include factorial design~\cite{raveendraparmar1987, guptaparmar1989,muruganParmar1994}, linear regression~\cite{yangbibby1993,kimsonyang2003}, multiple regression analysis~\cite{rajakumarmuralidharan2011}, response surface methodology~\cite{muruganparmar1997}, and Taguchi method~\cite{juangtarng2002,leehanhong2006}.
\subsection{Artificial Neural Networks and Automated Welding}
\label{ann}
Research on Artificial Neural Network (ANN) has witnessed a tremendous success in the past two decades. These computational models have been deployed in a wide range of applications from natural language processing~\cite{CollobertWeston2008} to object tracking~\cite{ThrunTracking2016}. In their industrial applications, automated welding is where these models have specially proven successful. In particular, the ability of these models to generalize on a broad range of inputs makes them an attractive choice, given the existence of a wide range of welding techniques~\footnote{Please visit $https://en.wikipedia.org/wiki/List\_of\_welding\_processes$ for comprehensive list of welding techniques.}.  In addition, ANN provides a powerful and accurate alternative to highly sophisticated yet domain-specific approaches such as heat flow~\cite{Rosenthal1941}, factorial design~\cite{kimsonyang2003}, response surface methodology~\cite{gunarajmurugan2005}, and support vector regression~\cite{SVR2, SVR1}.\newline
Application of ANN in estimation of the weld bead parameters bring a remarkable progress in recent years. These models are successfully deployed in investigation of the modeling, the process control, and the estimation of the quality of the weld beads~\cite{cookbarnett1995}. Anderson et al.~\cite{andersonramaswamy1990} investigate the application of multi-layer neural networks in modeling of the arc welding process. They achieve satisfactory results in their prediction accuracy as compared to the conventional control systems. Tang et al.~\cite{juangtarnglii1998} apply back-propagation and counter-propagation to analyze the relationship between IWP and the geometry of the weld bead in TIG welding. Polte et al.~\cite{WuPolte2000} consider the application of the self-organizing maps (SOM) in monitoring and quality evaluation of the GMAW welding. Nagesh and Datta~\cite{NageshDatta2002} apply back-propagation on modeling and estimation of the weld bead parameters in SMAW. Chandrasekhar and Vasudevan~\cite{chandrasekharvasudevan2010} combine their neural network with genetic algorithm for post-optimization of the estimated parameters of their network. Their model achieves a considerable improvement in optimization of these parameters in A-TIG welding. Benyounisa and Olabi~\cite{BenyounisaOlabi2008} present a comprehensive survey on optimization techniques for estimation of the direct weld parameters.
\subsection{Challenges and Limitations}
\label{limits}
One challenging aspect of this estimation problem is the dependency of DWP on the same input control signals or IWP. For instance, the same set of values of voltage, current, and torch traveling speed (i.e., IWP) produces the different values that pertain to the depth of penetration and the width of weld beads (i.e., DWP). Therefore, it is of no value to combine these parameters in a single output vector to estimate their values. This is mainly due to the pseudo-redundancy caused by the overlapping values of input feature vectors for different output data. This is particularly problematic when applying backpropagation~\cite{LeCunReview1998} due to the saturation behavior of the common activation functions in neural networks~\cite{GulcehreyBengio2016}. Utilization of the rare features to craft domain-specific feature weightings, such as TF-IDF~\cite{SaltonBuckley1988} is a solution to tame this limitation. However, the inherent low-dimensionality of IWP feature vectors (with voltage, current, and torch traveling speed among the most common values) does not guarantee the applicability of this solution in the domain of welding. This limits the ability of estimation models to establish a well-generalized mapping of different DWP onto the same set of IWP values. Although some models attempt to compensate for such shortcomings through introduction of polynomial features, such polynomial degrees do not necessarily guarantee a better estimation of DWP values~\cite{muruganParmar1994,gunarajmurugan2005}.\newline
On the other hand, adaptive gradient or \emph{adagrad} dynamically incorporates knowledge of the geometry of data that is observed in earlier iterations in order to achieve more informative gradient-based learning~\cite{adagrad2011}. However, the weak correlation among DWP (please refer to Figure~\ref{Penetration_Width_Correlation}) does not allow the inference of the value of one (e.g., depth of penetration) from the other (e.g., width).  Furthermore, some approaches apply noise to the activation function~\cite{Bengioetal2013, NairHinton2010} for feed-forward networks and Boltzmann machines~\cite{HintonBoltzmannMachine2012} to encourage deeper exploration by neural units, thereby easing the optimization of their corresponding outcomes. However, these approaches face vanishing gradient information due to the saturation of these units~\cite{GulcehreyBengio2016}. Although \emph{rmsprop}~\cite{rmsprop} shows impressive results and improvements~\cite{rmspropschaul2013}, there is not much understanding for its success in practice~\cite{DauphinBengio2015}. Momentum is another approach to increase speed of convergence. It damps the size of gradient steps along the directions of high curvature while yielding a larger effective learning rate along the directions of low curvature~\cite{momentum1}. As a result, it ensures faster convergence~\cite{nesterov2009,rouxmomentum2012} compared to stochastic gradient descent in convex optimization problems~\cite{GulcehreyBengio2016}. Thorough investigation of adopted techniques for backpropagation and gradient based training are found in~\cite{LeCunReview1998,BengioReview2012}.
\subsection{Contributions}
\label{contributions}
Our contributions to address these challenging issues are: 
\begin{enumerate}
\item{We resolve the vanishing gradient of our model through introduction of a new meta-parameter, referred to as \emph{reinforced learning coefficient}, to enable this model to maintain the quality of its computed gradients. Our approach is comparable to Gulcehre et. al~\cite{GulcehreyBengio2016}. Whereas, they apply an additive noise to the input or output layers, the multiplicative nature of our parameter that is directly applied to the calculated gradients magnifies the exponent of the activation function in next iteration, thereby sharpening its transition. In addition, we model a minimalist single-layer neural regression model as opposed to a deep neural network for classification in their study.}
\item{We further reduce the estimation error of our model using a weighted regression estimator at its output layer and learning the state of this weight.} 
\item{we show the scalability of our model on combination of datasets of different welding techniques. In particular, we show that the change in estimation error exhibited by our model due to combination of data of these welding techniques is statistically insignificant. This is a non-trivial result that challenges the general belief in this field of research. However, we acknowledge that our result is, by no means, a definitive solution but an early attempt to formulation of one such scalable model.}
\end{enumerate}
It is worth noting that we tame the effect of low-dimensionality of IWP by readjusting the number of neurons of the hidden layer of the respective blocks of our model at runtime and in consecutive training epochs. This enables our model to better capture the interrelation between individual elements of the input feature vectors (i.e., IWP), thereby eliminating the needs for further polynomial features~\cite{muruganParmar1994,gunarajmurugan2005}. Additionally, we present a comprehensive comparative study of our model in contrast to state-of-the-art techniques in the literature to show a significant improvement in estimation of direct weld parameters.\newline
The remainder of this article is organized as follows. Section~\ref{neuronArc} explains the overall architecture and the formulation of our neural regression network. We present the results of performance of our model in Section~\ref{cases}. In section~\ref{further1}, we study the effect of increase in hidden layers on estimation quality of our model. Section~\ref{further2} elaborates on scalability of our model on combined data of different welding techniques. Conclusion and some insights on future direction of this research are presented in Section~\ref{conclude}.
\section{Neural Architecture}
\label{neuronArc}
\begin{figure}[b!]
\begin{center}
\centerline{\includegraphics[height=1.3in,width=\columnwidth]{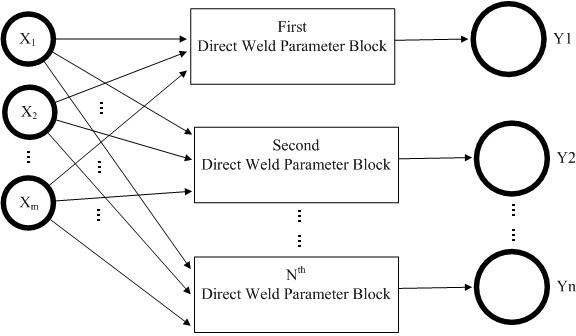}}
\caption{Overall Architecture. Each block is a one-layer neural regression network whose set of meta-parameters and direct weld parameter are independent of other blocks. IWP form the input features to the entire model.}
\label{neuralArc}
\end{center}
\end{figure} 

\begin{figure}[t!]
\begin{center}
\centerline{\includegraphics[height=1.4in,width=\columnwidth]{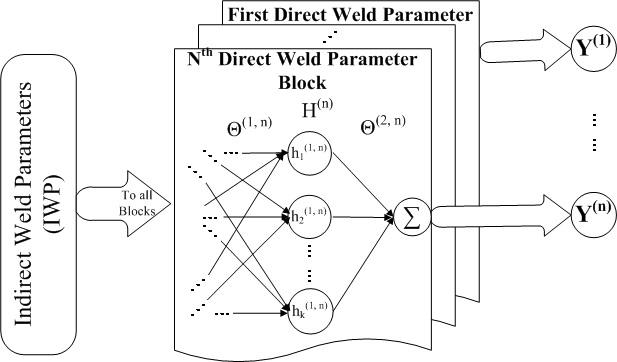}}
\caption{Block-level neural architecture. \emph{n} is the total number of blocks (i.e., cardinality of DWP) and \emph{k} indicates the number of neurons on a hidden layer. The weight matrices \emph{$\Theta^{(i,n)},~i=1, 2$} are unique to their blocks.}
\label{BlockLevelModel}
\end{center}
\end{figure}

We model our minimalist neural architecture as an aggregate of independent one-hidden-layer regression blocks as depicted in Figure~\ref{neuralArc}. Number of these blocks is proportional to cardinality of DWP. For instance, it is a two-block regression model if  DWP are depth of penetration and width of weld beads. These blocks are trained simultaneously and share an identical structure as shown in Figure~\ref{BlockLevelModel}. In particular, they apply Sigmoid activation function on IWP feature vectors and generate their corresponding outputs through application of linear transformation function on calculated values of their respective hidden layer. This enables each block to act as an independent non-linear regression network. Initially, the hidden layer of a given block is assigned with two neural computation units. During the training, our model modifies the size of this hidden layer if such an update results in further improvement of their respective estimated values.\newline
Let \emph{$X$} denote the set of feature vectors (i.e., IWP) to our model. Furthermore, let \emph{$Y^{(k)}$} and \emph{$H^{(k)}$} represent the set of direct weld parameter and the hidden layer of the \emph{$k^{th}$} block of this architecture, with \emph{$k=1,\dots, N$}, where \emph{N} is the cardinality of DWP. Moreover, let \emph{$\Theta^{(r,k)},~r=1, 2$}, be the weight matrices that are associated with the \emph{$k^{th}$} block. Values that are generated at the hidden layer of the \emph{$k^{th}$} block are:
\begin{align}
\begin{split}\label{zValues}
z^{(1,k)} = X^T \times \Theta^{(1,k)}
\end{split}\\
\begin{split}\label{hiddenValues}
H^{(k)} = \frac{1}{1 + e^{-z^{(1,k)}}},~k=1,\dots, N
\end{split}
\end{align}
where \emph{N} is the cardinality of DWP. The final \emph{weighted estimates} of these blocks at \emph{$t^{th}$} iteration of the training are:
\begin{equation}
Y'^{(k, t)} =[(H^{(k)})^T \times \Theta^{(2,k)}] + \tau^{(k, t)}
\label{finalOutput}
\end{equation}
\emph{$\Theta^{(2,k)}$} is the weight matrix between the hidden layer of the \emph{$k^{th}$} block and its output layer. \emph{$\tau^{(k, t)}$} in equation~(\ref{finalOutput}) is the weight associated with the regression estimator at the output layer of the \emph{$k^{th}$} block whose value is learned during the training phase:
\begin{eqnarray}
\tau^{(k, t)} = \{-\nu, 0, \nu\}\\
\nu = \frac{1}{m}(Y'^{(k, t-1)} - Y^{(k)})
\label{WeightedOutput}
\end{eqnarray}
subject to:
\begin{equation}
\min [\frac{1}{2m}[ (Y^{(k)} - Y'^{(k, t)})^2 + \lambda^{(k)} \sum_{r=1}^{2} \sum_{i=1}^{p^{(r,k)}} \sum_{j=1}^{q^{(r,k)}} (\Theta^{(r,k)}_{ji})^2]]
\label{costFunc}
\end{equation} 
In other words, the value of this weight varies between zero and \emph{$\pm$} mean of actual estimation errors at \emph{$t-1$} iteration of the training, given its positive impact on minimization of the regularized cost function of its respective regression block as formulated in equation~(\ref{costFunc}). The entries of \emph{$Y'^{(k, t-1)}$} in equation~(\ref{WeightedOutput}) are initialized to zero at \emph{$t=1$} and \emph{m} is the total number of training data. The second term in equation~(\ref{costFunc}) is the regularization term that incorporates the sum of the squared of the weight matrices in the \emph{$k^{th}$} block to prevent overfitting on its corresponding direct weld parameter. \emph{$\lambda^{(k)}$} is its regularization factor. \emph{$p^{(r,k)}$} and \emph{$q^{(r,k)},~r=1, 2$} refer to the row and the column dimensions of the weight matrices of the \emph{$k^{th}$} block, respectively. It is worth noting that we add an extra column with all whose entries equals to 1 to \emph{X} and \emph{$H^{(k)}$} to count for the bias term. This is why we do not 
explicitly include this term in equation~(\ref{zValues}). This also explains the starting indices of the last two summation operations from 1 in equation~(\ref{costFunc}).\newline 
The discrepancies between these \emph{weighted estimates} in equation~(\ref{finalOutput}) and the actual DWP at \emph{$t^{th}$} training iteration are:
\begin{equation}
\delta^{(Y^{(k)})} = Y^{(k)} - Y'^{(k, t)}, ~k=1,\dots, N
\label{deltaOutput}
\end{equation} 
We use these values to update \emph{$\Theta^{(2,k)}$}:
\begin{align}
\begin{split}
\label{theta5_1}
\Delta^{(2,k)} = ((\delta^{(Y^{(k)})})^T \times H^{(k)}) \times \gamma^{(k)}
\end{split}\\
\begin{split}
\label{theta5_2}
\Theta^{(2,k)} = \Theta^{(2,k)} + \frac{1}{m}(\alpha^{(k)} \times \Delta^{(2,k)} + \lambda^{(k)} \sum_{i=1}^{p^{(2,k)}}\sum_{j=1}^{q^{(2,k)}}\Theta^{(2,k)}_{ji})
\end{split}
\end{align}
where \emph{m} is the total number of training data and \emph{$k=1,\dots, N$} reflects the total number of blocks (i.e., cardinality of DWP) in our model. \emph{$p^{(2,k)}$} and \emph{$q^{(2,k)}$} refer to the dimensions of the weight matrix of the linear transformation layer of \emph{$k^{th}$} block.  Equation~(\ref{theta5_1}) is the gradient of this output layer and \emph{$\gamma^{(k)}$} is the \emph{reinforced learning coefficient} of the \emph{$k^{th}$} block. In essence, \emph{$\gamma^{(k)}$} is a standard normal multiplicative factor (i.e., \emph{$((\sigma^2 \times \gamma^{(k)}) + \mu) \sim \mathcal{N}(\mu=0, \sigma^2=1)$}) that is learned during the training and is shared by all the layers of a given block. It magnifies the exponent of the Sigmoid function in each consecutive iteration via perturbing the gradients of the previous step during the training, thereby maintaining stronger gradients.This allows our model to further reduce its deviation from the expected DWP to significantly minimize its estimation error. Furthermore, \emph{$\alpha^{(k)}$} and \emph{$\lambda^{(k)}$} are the learning rate and the regularization factor of the corresponding independent set of meta-parameters of \emph{$k^{th}$} block, respectively. Similarly, the update of the weight matrix \emph{$\Theta^{(1,k)}$} is:
\begin{align}
\begin{split}\label{theta432_1}
\delta^{(2,k)}  = ((\Theta^{(2,k)})^T \times \delta^{(Y^{(k)})}) \times (H^{(k)} \times (1 - H^{(k)}))\\,~k=1,\dots, N
\end{split}\\
\begin{split}\label{theta432_2}
\Delta^{(1,k)} = ((\delta^{(2,k)})^T \times X) \times \gamma^{(k)}
\end{split}\\
\begin{split}\label{theta432_3}
\Theta^{(1,k)} = \Theta^{(1,k)} + \frac{1}{m}(\alpha^{(k)} \times \Delta^{(1,k)} + \lambda^{(k)} \sum_{i=1}^{p^{(1, k)}}\sum_{j=1}^{q^{(1,k)}}\Theta^{(1,k)}_{ji})
\end{split}
\end{align}
where \emph{$(H^{(k)} \times (1 - H^{(k)}))$} in equation~(\ref{theta432_1}) is the gradient of the Sigmoid activation function of the \emph{$k^{th}$} block.
\section{Case Study}
\label{cases}
\begin{figure}[b!]
\centerline{\includegraphics[height=1.7in, width=3.2in]{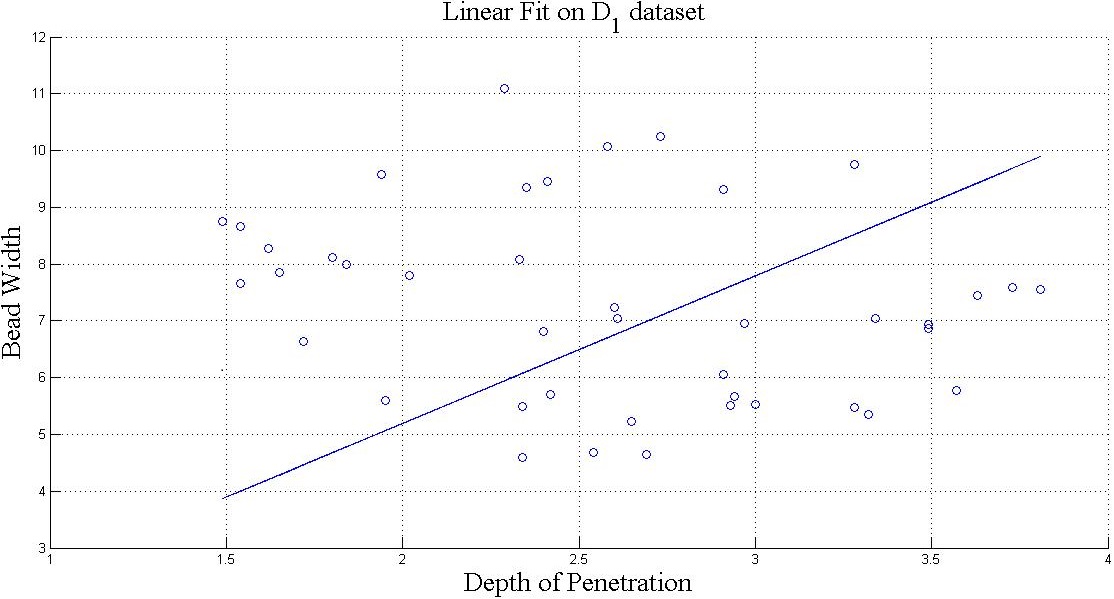}}
\caption{Regression fit on $D_1$~\cite{andersonramaswamy1990}, demonstrating a weak correlation between depth of penetration and width of weld beads.\vspace{-5mm}} 
\label{Penetration_Width_Correlation}
\end{figure}

\begin{figure}[b!]
\centerline{\includegraphics[height=2.0in, width=3.2in]{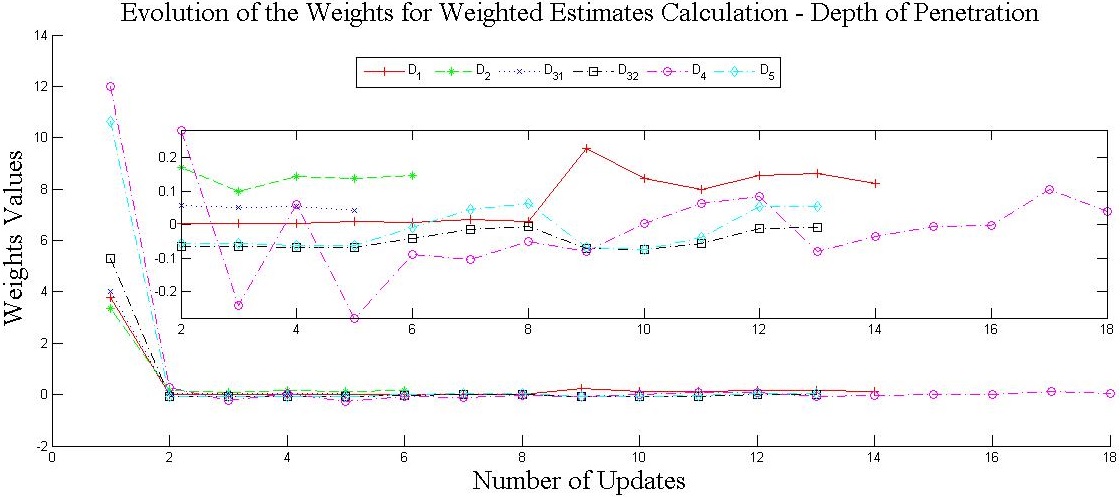}}
\caption{Evolution of weight values i.e., \emph{$\tau^{(k, t)}$} in equation~(\ref{finalOutput}) of weighted estimates for Depth of Penetration regression block on datasets \emph{$D_1$} through \emph{$D_{5}$}.} 
\label{bestOffsets_penetration}
\end{figure}

\begin{figure}[b!]
\centerline{\includegraphics[height=2.0in, width=3.2in]{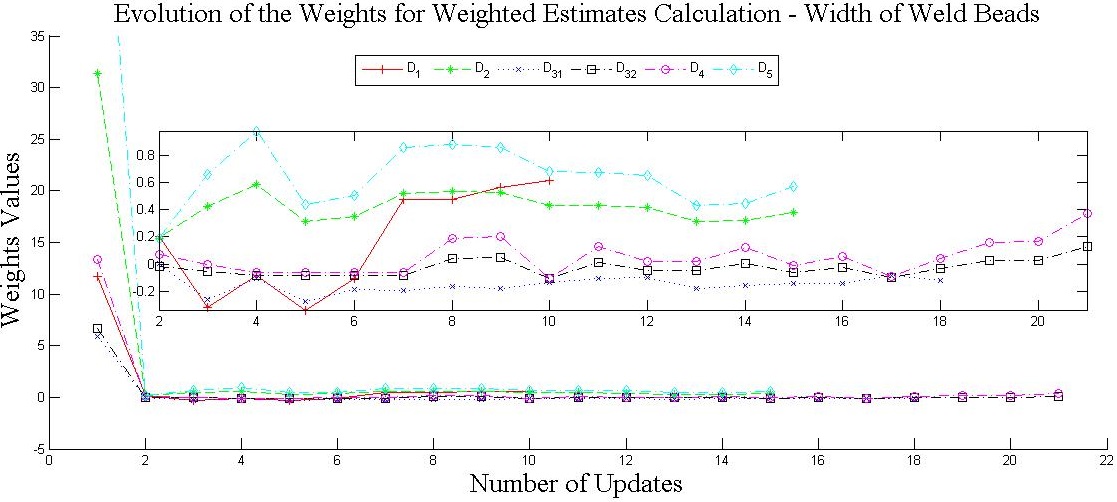}}
\caption{Evolution of weight values i.e., \emph{$\tau^{(k, t)}$} in equation~(\ref{finalOutput}) of weighted estimates for Width of Weld Beads regression block on datasets \emph{$D_1$} through \emph{$D_{5}$}.} 
\label{bestOffsets_width}
\end{figure}

We compare the performance of our minimalist neural regression network in contrast to a number of selected studies in the literature. In particular, we choose Anderson et al.~\cite{andersonramaswamy1990}, Chandrasekhar and Vasudevan~\cite{chandrasekharvasudevan2010}, Karaoglu and Secgin~\cite{karaoglusecgin2008}, Chotěborský et.al~\cite{FactorialGuys2011}, and Dhas and Satheesh~\cite{dhassatheesh2013}. These articles form a reliable representatives of the trends of research in this field over the past two and a half decades. More specifically,~\cite{andersonramaswamy1990} is the seminal paper that introduces the application of multi-layer neural networks in estimation of the direct weld parameters. Furthermore, Karaoglu and Secgin~\cite{karaoglusecgin2008} adopt a curvilinear approach to estimation of DWP. Moreover, Chandrasekhar and Vasudevan~\cite{chandrasekharvasudevan2010} propose a combination of single hidden layer neural network and genetic algorithm for post-optimization of the estimated DWP. In addition, Chotěborský et.al~\cite{FactorialGuys2011} and Dhas and Satheesh~\cite{dhassatheesh2013} present (to the best of our knowledge) the highest estimation accuracy in application of the classical models, Taguchi method and factorial design respectively, in the literature. Furthermore, their welding data that are publicly available, pertain to four main welding techniques, namely:
\begin{enumerate}
\item{\emph{$D_1$}: Anderson et al.~\cite{andersonramaswamy1990} on arc welding process.}
\item{\emph{$D_2$}:  Karaoglu and Secgin~\cite{karaoglusecgin2008} on submerge arc welding (SAW).}
\item{\emph{$D_{31}$} and \emph{$D_{32}$}: Chandrasekhar and Vasudevan~\cite{chandrasekharvasudevan2010} datasets on Tungsten insert gas welding (A-TIG) welding on 304LN and 316LN stainless steel, respectively.}
\item{\emph{$D_4$}: Dhas and Satheesh~\cite{dhassatheesh2013} dataset on Taguchi method on SAW.}
\item{\emph{$D_5$}: Chotěborský et.al~\cite{FactorialGuys2011} dataset on factorial design technique on gas metal arc welding (GMAW).}
\end{enumerate}
Figure~\ref{Penetration_Width_Correlation} shows the regression fit to the depth of penetration and width of weld beads in \emph{$D_1$}. Their Pearson correlation ratio along with its p-value are \emph{$\{0.12, -0.24\}$}. Furthermore, those of Spearman and Kendall for these DWP are \emph{$\{0.05, -0.31\}$} and \emph{$\{0.09, -0.17\}$}, respectively. These correlation ratios show a weak proportionality between the values of these DWP. Furthermore, their p-values reject the null hypothesis of correlation between these DWP at \emph{$0.05$} and \emph{$0.01$} significance levels. This is in accordance with our claim in section~\ref{limits} of this article. Moreover, this trend of  absence of correlation between DWP is exhibited in \emph{$D_2$} through \emph{$D_{5}$}.\newline
Tables~\ref{meta1} and~\ref{meta2} reveal the overall simplicity of our minimalist neural regression model. More specifically, the cumulative number of neurons of the two blocks of our model that are dynamically selected at runtime from range \emph{$\{2,\dots,100\}$}, does not exceed
\begin{table}[b!]
\begin{center}
{\caption{Best meta-parameters - depth of penetration block.}
\label{meta1}}
\begin{tabular}{lcccccc}
\hline
\rule{0pt}{12pt}
Datasets&\parbox[t]{1.0cm}{Neurons $\in [2,\dots\\, 100]$}&Degree&$\alpha$&$\gamma$&$\lambda$&\parbox[t]{1.0cm}{Iterations\\ $\in [1000,\dots\\, 12000]$}\\
\hline
\\[-6pt]
\quad $D_1$& 4& 0& 40.0 &0.9& 0.0 & 1000\\
\quad$D_2$& 8& 0 & 100 & 0.1 & 0.001 & 1000\\
\quad $D_{31}$& 5& 0 & 30 & 1.0 & 0.0& 9000\\
\quad $D_{32}$&9 & 0 &  0.009 & 2500 & 0.0 & 7000\\
\hline
\end{tabular}
\end{center}
\end{table}
\begin{table}[t!]
\begin{center}
{\caption{Best meta-parameters - width of weld beads block.}
\label{meta2}}
\begin{tabular}{lcccccc}
\hline
\rule{0pt}{12pt}
Datasets&\parbox[t]{1.0cm}{Neurons $\in [2,\dots\\, 100]$}&Degree&$\alpha$&$\gamma$&$\lambda$&\parbox[t]{1.0cm}{Iterations\\ $\in [1000,\dots\\, 12000]$}\\
\hline
\\[-6pt]
\quad $D_1$& 2 & 0& 0.01 & 2500 & 0.0 & 9000\\
\quad$D_2$&3 & 0 & 0.009 & 2000 & 0.003 & 1000\\
\quad $D_{31}$&4 & 0 & 0.3 & 1500 & 0.0 & 1000\\
\quad $D_{32}$&3 & 0 & 50 & 0.5 & 0.001 & 1000\\
\hline
\end{tabular}
\end{center}
\end{table}
\begin{figure}[b!]
\centerline{\includegraphics[height=1.7in, width=3.2in]{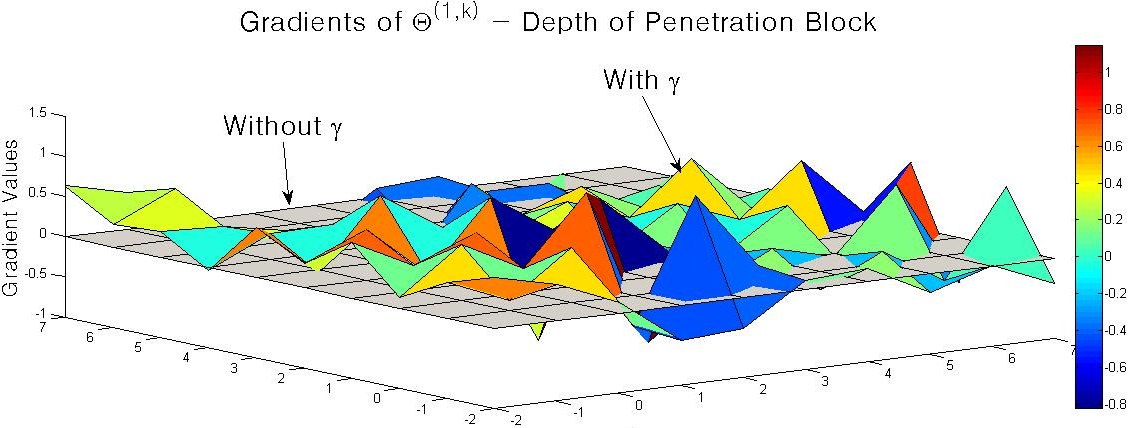}}
\caption{Effect of the Reinforced Learning Coefficient $\gamma$ on gradients of the weight matrix $\Theta^{(1, k)},~k=1$ i.e., depth of penetration block.} 
\label{GradientT1}
\end{figure}
\begin{figure}[t!]
\centerline{\includegraphics[height=1.2in]{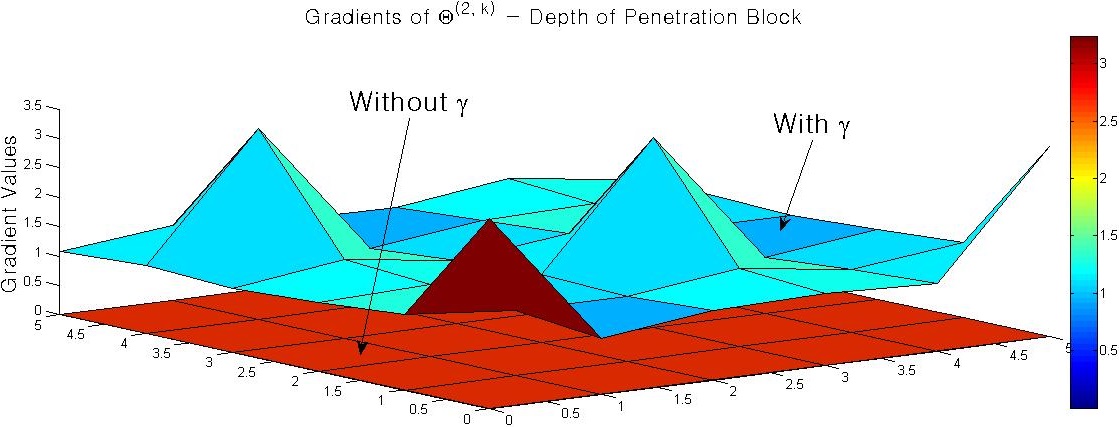}}
\caption{Effect of the Reinforced Learning Coefficient $\gamma$ on gradients of the weight matrix $\Theta^{(2, k)},~k=1$ i.e., depth of penetration block.} 
\label{GradientT2}
\end{figure}
 12 neurons in total (with mean and standard deviation of 9.50 and 2.65). It is interesting to note that the best results reported in~\cite{andersonramaswamy1990} correspond to a \emph{$2 \times 18$} networks (i.e., 36 neurons in total). Moreover, our model does not require any polynomial features (examined in the range \emph{$\{0,\dots, 6\}$}). This considerably reduces its training time. Additionally, it does not suffer from overfitting as indicated by significantly small entries of regularization parameter \emph{$\lambda$} in these tables, with most of whose entries equal zero.\newline
Moreover, these tables reveal the inverse proportionality of the values of the learning rate \emph{$\alpha$} and the \emph{reinforced learning coefficient} \emph{$\gamma$}. This implies that the readjustment of the gradients of the weight matrices of our model results in better compensation for the variance that is exhibited by the input feature vectors (i.e., IWP). Figures~\ref{GradientT1} and~\ref{GradientT2} illustrate the effect of the reinforced learning coefficient \emph{$\gamma$} on gradients of the weight matrices \emph{$\Theta^{(i,k)},~i,=1,~2,~~k=1$} i.e., depth of penetration block of our model. Furthermore, the evolution of the learned weight values i.e., \emph{$\tau^{(k, t)}$} in equation~(\ref{finalOutput}) for depth of penetration and the width of weld beads blocks of our model are shown in figures~\ref{bestOffsets_penetration} and~\ref{bestOffsets_width}, respectively.\newline
However, variations in number of iterations during the training (with mean and standard deviation of (4500, 4123.1) and (3000, 4000.0) for depth of penetration and width of weld beads, respectively) do not exhibit any particular pattern or correspondence with the size of given datasets or the DWP. As a result, any conclusion on the effect of the size of the dataset or type of direct weld parameter on training process is not warranted. However, it takes longer, on average, for our model to converge on depth of penetration as compared to width of weld beads.

\subsection{Comparison with Selected Literature}
Tables~\ref{t1990} through~\ref{newTable_2} show the results of the performance of our minimalist neural regression network (NRN in these tables) on test data in comparison with results reported in \emph{$D_1$}, \emph{$D_2$}, \emph{$D_{31}$}, \emph{$D_{32}$}, \emph{$D_4$}, and \emph{$D_5$}. In addition to these results, we also report the performance of our neural regression network in contrast with support vector regression (SVR)~\cite{SVR2, SVR1}, and neural network without application of our proposed \emph{reinforced learning coefficient} with \emph{weighted estimates} in equations~(\ref{finalOutput}),~(\ref{theta5_1}), and~(\ref{theta432_2}), (i.e., ANN entry of these tables). It is worth noting that we use the Python scikit-learn package~\cite{pythonpkg} for modelling SVR. It achieves its best results using the polynomial degree activation function (i.e., \emph{poly}) with four added polynomial features. Moreover, we choose Root Mean Squared Error (RMSE) and the percentage of Prediction Error (PE) to report our results. We calculate the PE of our model as:
\begin{align}
\begin{split}\label{PEs}
PE = \frac{|Y^{(k)} - Y'^{(k,T)}|}{Y^{(k)}} \times 100
\end{split}
\end{align}
where \emph{$Y^{(k)}$} and \emph{$Y'^{(k,T)}$} are the actual and estimated direct weld parameter of the \emph{$k^{th}$} block and \emph{$|.|$} gives the absolute value of its argument. \emph{T} in \emph{$Y'^{(k,T)}$} indicates the final estimated value of the \emph{$k^{th}$} block at time \emph{T}. We choose RMSE and PE to allow the comparative analysis of our results with respect to the selected literature. More specifically, Anderson et al.~\cite{andersonramaswamy1990} and Dhas and Satheesh~\cite{dhassatheesh2013} report the PE of their models. Whereas, Karaoglu and Secgin~\cite{karaoglusecgin2008} and Chandrasekhar and Vasudevan~\cite{chandrasekharvasudevan2010} provide RMSE as a measure of accuracy of their models. Furthermore, Chotěborský et.al~\cite{FactorialGuys2011} report Mean Squared Error (MSE) of their model. However, we include square root of these values i.e., RMSEs of their model.\newline
A comparison between the PE values in Table~\ref{t1990} reveals that the differences between these values are above one
\begin{table}[b!]
\begin{center}
{
\caption{RMSEs and PEs - neural regression network (NRN) vs. $D_1$, support vector regression (SVR), and regular artificial neural network (ANN).}
\label{t1990}}
\begin{tabular}{lcccccc}
\hline
\rule{0pt}{12pt}
&Penetration (mm)&\multicolumn{3}{c}{Width}\\
\cline{1-5}
\rule{0pt}{12pt}
&$RMSE$&$PE$&$RMSE$&$PE$\\
\hline
\\[-6pt]
\quad$D_1$&-&19.58\%&-&5.68\%\\
\quad NRN&0.085&3.59\%& 0.073&2.91\%\\
\quad SVR&0.34&25.80\%&0.26&23.13\%\\
\quad ANN&0.17&11.62\%&0.28&24.61\%\\
\hline
\end{tabular}
\end{center}
\end{table}
 standard deviation (9.65 and 11.37, respectively). As a result, the reduction of the percentage of the prediction error by our model on this dataset is significant. Additionally, the RMSE values that are calculated by our model are considerably small (in the scale of millimeter). However, we are unable to confirm the significance of these values since the RMSEs of the multi-layer network in~\cite{andersonramaswamy1990} are not reported.\newline
We notice the same trend of improvement on the entries of Tables~\ref{t2008} through~\ref{t2010_2}. More specifically, Table~\ref{t2008} shows that our model achieves more than 8 times improvements on RMSE values compared to \emph{$D_2$}, with its corresponding \emph{z-score} above one standard deviation (it is 1.382 and -1.457 for depth of penetration and width of weld beads, respectively). Moreover, the PE values reported in this table are significantly small with whose difference above one standard deviation from those of \emph{SVR} and \emph{ANN} (6.801 and 5.596 for depth of penetration and width of weld beads). However, we cannot draw any conclusion on its significance with respect to~\cite{karaoglusecgin2008} since 
\begin{table}[t!]
\begin{center}
{
\caption{RMSEs and PEs - neural regression network (NRN) vs. $D_2$, support vector regression (SVR), and regular artificial neural network (ANN).}
\label{t2008}}
\begin{tabular}{lcccccc}
\hline
\rule{0pt}{12pt}
&Penetration (mm)&\multicolumn{3}{c}{Width}\\
\cline{1-5}
\rule{0pt}{12pt}
&$RMSE$&$PE$&$RMSE$&$PE$\\
\hline
\\[-6pt]
\quad$D_2$&0.292&-&0.353&-\\
\quad NRN&0.034&0.85\%& 0.041&0.94\%\\
\quad SVR&0.14&10.18\%&0.11&9.13\%\\
\quad ANN&0.06&2.75\%&0.16&7.43\%\\
\hline
\end{tabular}
\end{center}
\end{table}
they do not provide PEs of their model. Similarly, Tables~\ref{t2010_1} and~\ref{t2010_2} report the improvements of the RMSE values that range between 1.62 - 3.01 and 3.35 - 3.81 times, compared to results obtained in \emph{$D_{31}$} and \emph{$D_{32}$}, respectively. Moreover, the PE values of these tables are above one standard deviation (i.e., (4.86, 4.24) and  (5.81, 5.09) for entries of Tables~\ref{t2010_1} and~\ref{t2010_2}, respectively). Additionally, these improvements are the direct results of our minimalist neural regression network using reinforced learning coefficient and weighted estimates, without application of any post-optimization of the estimated values using genetic algorithm, as reported by Chandrasekhar and Vasudevan~\cite{chandrasekharvasudevan2010}. Additionally, our model achieves 1.40 and 4.28 times improvements on depth of penetration and width of weld beads in \emph{$D_4$} , with their difference above one standard deviation (it is 2.340) on width of weld beads, as shown in Table~\ref{newTable_1}. Furthermore, the \emph{z-score} of its RMSEs are above one standard deviation from those calculated by SVR and ANN (they are -1.154 and -1.126, respectively). However, we are unable to report the significance of these RMSEs with respect to results in \emph{$D_4$} since Dhas and Satheesh~\cite{dhassatheesh2013} do not present these values on their model. Similarly our model outperforms results by Chotěborský et.al~\cite{FactorialGuys2011} in \emph{$D_5$} by a large margin i.e., 9.5 and 15 times improvements on DWP, as shown in Table~\ref{newTable_2}. \newline
Moreover, the overlapping interval at $95\%$ confidence level in Table~\ref{memyself} indicates that differences between the results obtained by our model on these datasets is insignificant. In other words, our minimalist model using \emph{reinforced learning coefficient} with \emph{weighted estimates} is capable of generalizing on features of different welding techniques. This observation is further supported by the minor differences in the meta parameters associated with these datasets on Tables~\ref{meta1} and~\ref{meta2}. 
\begin{table}[b!]
\begin{center}
{
\caption{RMSEs and PEs - neural regression network (NRN) vs. $D_{31}$, support vector regression (SVR), and regular artificial neural network (ANN).}
\label{t2010_1}}
\begin{tabular}{lcccccc}
\hline
\rule{0pt}{12pt}
&Penetration (mm)&\multicolumn{3}{c}{Width}\\
\cline{1-5}
\rule{0pt}{12pt}
&$RMSE$&$PE$&$RMSE$&$PE$\\
\hline
\\[-6pt]
\quad$D_{31}$&0.148&-&0.205&-\\
\quad NRN&0.091&2.65\%& 0.068&1.94\%\\
\quad SVR&0.299&10.18\%&0.159&9.13\%\\
\quad ANN&0.263&11.75\%&0.165&9.43\%\\
\hline
\end{tabular}
\end{center}
\end{table}

\begin{table}
\begin{center}
{
\caption{RMSEs and PEs - neural regression network (NRN) vs. $D_{32}$, support vector regression (SVR), and regular artificial neural network (ANN).}
\label{t2010_2}}
\begin{tabular}{lcccccc}
\hline
\rule{0pt}{12pt}
&Penetration (mm)&\multicolumn{3}{c}{Width}\\
\cline{1-5}
\rule{0pt}{12pt}
&$RMSE$&$PE$&$RMSE$&$PE$\\
\hline
\\[-6pt]
\quad$D_{32}$&0.124&-&0.156&-\\
\quad NRN&0.037&0.93\%& 0.041&1.04\%\\
\quad SVR&0.246&13.48\%&0.193&11.66\%\\
\quad ANN&0.226&11.75\%&0.165&9.43\%\\
\hline
\end{tabular}
\end{center}
\end{table}

\begin{table}
\begin{center}
{
\caption{RMSEs and PEs - neural regression network (NRN) vs. $D_{4}$, support vector regression (SVR), and regular artificial neural network (ANN).}
\label{newTable_1}}
\begin{tabular}{lcccccc}
\hline
\rule{0pt}{12pt}
&Penetration (mm)&\multicolumn{3}{c}{Width}\\
\cline{1-5}
\rule{0pt}{12pt}
&$RMSE$&$PE$&$RMSE$&$PE$\\
\hline
\\[-6pt]
\quad$D_{4}$&-&1.27\%&-&4.32\%\\
\quad NRN&0.026&0.91\%& 0.037&1.01\%\\
\quad SVR&0.258&14.48\%&0.212&12.66\%\\
\quad ANN&0.267&10.75\%&0.172&8.94\%\\
\hline
\end{tabular}
\end{center}
\end{table}
\begin{table}[t!]
\begin{center}
{
\caption{RMSEs and PEs - neural regression network (NRN) vs. $D_5$, support vector regression (SVR), and regular artificial neural network (ANN).}
\label{newTable_2}}
\begin{tabular}{lcccccc}
\hline
\rule{0pt}{12pt}
&Penetration (mm)&\multicolumn{3}{c}{Width}\\
\cline{1-5}
\rule{0pt}{12pt}
&$RMSE$&$PE$&$RMSE$&$PE$\\
\hline
\\[-6pt]
\quad$D_5$&0.228&-&0.570&-\\
\quad NRN&0.024&0.75\%& 0.038&0.84\%\\
\quad SVR&0.18&10.28\%&0.15&9.64\%\\
\quad ANN&0.069&3.01\%&0.172&8.12\%\\
\hline
\end{tabular}
\end{center}
\end{table}

\begin{table}
\begin{center}
{
\caption{RMSE Confidence Interval (95\%) - minimalist neural regression network (NRN) on $D_1$, $D_2$, $D_{31}$, $D_{32}$, $D_{4}$, and $D_{5}$.}
\label{memyself}}
\begin{tabular}{lcccc}
\hline
\rule{0pt}{12pt}
&$Penetration$&$Width$\\
\hline
\\[-6pt]
\quad $D_1$&(0.019, 0.143)&(0.031, 0.155)\\
\quad $D_2$&(-0.031, 0.102)&(-0.040, 0.084)\\
\quad $D_{31}$&(0.051, 0.175)&(-0.008, 0.120)\\
\quad $D_{32}$&(-0.019, 0.110)&(-0.006, 0.118)\\
\quad $D_{4}$&(-0.017, 0.108)&(-0.004, 0.113)\\
\quad $D_{5}$&(-0.015, 0.098)&(-0.004, 0.114)\\
\hline
\end{tabular}
\end{center}
\end{table}

\subsection{Comparison with Adagrad, Rmsprop, and Nesterov Momentum}
\label{compare}
We compare the performance of our minimalist model using \emph{reinforced learning coefficient} with \emph{weighted estimates} in equations~(\ref{finalOutput}), (\ref{theta5_1}), and~(\ref{theta432_2}) in contrast with neural network using \emph{adagrad}~\cite{adagrad2011}, \emph{rmsprop}~\cite{rmsprop}, and \emph{Nesterov momentum}~\cite{nesterov2009} for backpropagation and weight updates. Increase in the average number of neural units is the first difference between our approach and these techniques. Whereas, number of neurons in our model does not exceed 12 units in total (with mean and standard deviation of 9.50 and 2.65) as shown in Tables~\ref{meta1} and~\ref{meta2}, the average number of neurons is more than 70 for a model using these alternative techniques. Furthermore, our approach achieves a considerable improvements on calculating RMSEs as compared to their results as shown in Tables~\ref{backpropPenet} and~\ref{backpropWidth}. In particular, such an improvement is significant when we compare our approach with \emph{Nesterov momentum} for both, depth of penetration and width of weld beads, with their difference above one standard deviation, on average (2.617 and 2.617, respectively). Moreover, performance of \emph{adagrad} and \emph{rmsprop} on these datasets are equivalent. The average \emph{z-score} of the performance of our approach is above one standard deviation of RMSEs computed by \emph{adagrad} and \emph{rmsprop} (their \emph{z-score} values are [-1.149 0.483 0.666] and [-1.130, 0.363, 0.767] for depth of penetration and width of weld beads for our model, \emph{adagrad} and \emph{rmsprop}, respectively). Additionally, our model achieves, on average, 2.817 and 2.871 times improvements on depth of penetration and 4.868 and 5.153 on width of weld beads. As a result, these improvements exhibited by our model are statistically significant.
\begin{table}[t!]
\begin{center}
{
\caption{RMSEs - neural regression network (NRN) vs. neural network using adagrad, rmsprop, and Nesterov Momentum.}
\label{backpropPenet}}
\begin{tabular}{lcccc}
\hline
\rule{0pt}{12pt}
&\multicolumn{4}{c}{Depth of Penetration (mm)}\\
\cline{1-5}
\rule{0pt}{12pt}
&$NRN$&$adagrad$&$rmsprop$&$Nesterov$\\
\hline
\\[-6pt]
\quad $D_1$&0.085&0.173&0.186&3.435\\
\quad $D_2$&0.034&0.091&0.091&3.261\\
\quad $D_{31}$&0.091&0.267&0.186&4.139\\
\quad $D_{32}$&0.037&0.125&0.224&3.593\\
\quad $D_{4}$&0.026&0.165&0.218&3.942\\
\quad $D_{5}$&0.024&0.138&0.233&3.867\\
\hline
\end{tabular}
\end{center}
\end{table}

\begin{table}[t!]
\begin{center}
{
\caption{RMSEs - neural regression network (NRN) vs. neural network using adagrad, rmsprop, and Nesterov Momentum.}
\label{backpropWidth}}
\begin{tabular}{lcccc}
\hline
\rule{0pt}{12pt}
&\multicolumn{4}{c}{Width of Weld Beads (mm)}\\
\cline{1-5}
\rule{0pt}{12pt}
&$NRN$&$adagrad$&$rmsprop$&$Nesterov$\\
\hline
\\[-6pt]
\quad $D_1$&0.073&0.362&0.368&3.630\\
\quad $D_2$&0.041&0.291&0.332&3.856\\
\quad $D_{31}$&0.068&0.352&0.341&4.026\\
\quad $D_{32}$&0.041&0.123&0.085&3.215\\
\quad $D_{4}$&0.037&0.146&0.223&4.108\\
\quad $D_{5}$&0.038&0.152&0.203&3.675\\
\hline
\end{tabular}
\end{center}
\end{table}

\begin{table}[b!]
\begin{center}
{
\caption{Neurons per hidden layer and the Iterations - Average performance of NRN with $1,\dots,~4$ hidden layers on $D_1$ through $D_{5}$.}
\label{netsALL}}
\begin{tabular}{lcccccc}
\hline
\rule{0pt}{12pt}
&Penetration (mm)&\multicolumn{3}{c}{Width}\\
\cline{1-5}
\rule{0pt}{12pt}
&$Nrns$&$Iters$&$Nrns$&$Iters$\\
\hline
\\[-6pt]
\quad$1H$&6.5&4500&3&3000\\
\quad$2H$&7&7000& 6&5000\\
\quad$3H$&9&7000&7&1000\\
\quad$4H$&9&6000&7&4000\\
\hline
\end{tabular}
\end{center}
\end{table}

\section{Effect of Number of Hidden Layers on Quality of Estimation}
\label{further1}
Table~\ref{netsALL} presents average number of neural units (per hidden layer) as well as average iterations for training our model on \emph{$D_1$} through \emph{$D_{5}$}, with its number of hidden layers varying in range \emph{$[1,\dots,4]$}. It is worth noting that our experiment with increased number of hidden layers indicates that the estimated values rapidly degrade, trapping in a zero-gradient regime, once the number of hidden layers exceeds four. Therefore, we do not report on architecture with more than four hidden layers.\newline
Table~\ref{netsALL} shows that the difference between the number of neurons of these architectures for depth of penetration and width of weld beads are above one standard deviation (1.315 and 1.893, respectively). Moreover, number of iterations to train these models proportionally increases with an increase in number of hidden layers, in case of depth of penetration. However, this observation does not hold in case of width of weld beads.\newline
A comparison between the RMSE and PE values of these models in Table~\ref{rmseALL} for depth of penetration and width of weld beads reveals that their corresponding values are within one standard deviation from each other (they are (0.003, 0.436) and (0.0024, 0.642) for RMSE and PE values). In addition, Table~\ref{allNets_ci} shows that the confidence interval of these models for \emph{$95\%$} level are all overlapping. As a result, an increase in the number of hidden layers of our minimalist neural regression network does not guarantee a significant improvement on its performance for estimation of DWP of the welding processes. Therefore, the minimalist design of our neural regression network does not jeopardize its quality of  estimation. 
\begin{table}[b!]
\begin{center}
{
\caption{RMSEs and PEs - Average performance of NRN with $1,\dots,~4$ hidden layers on $D_1$ through $D_{5}$.}
\label{rmseALL}}
\begin{tabular}{lcccccc}
\hline
\rule{0pt}{12pt}
&Penetration (mm)&\multicolumn{3}{c}{Width}\\
\cline{1-5}
\rule{0pt}{12pt}
&$RMSE$&$PE$&$RMSE$&$PE$\\
\hline
\\[-6pt]
\quad$1H$&0.061&2.005\%&0.055&1.707\%\\
\quad$2H$&0.069&2.98\%& 0.052&2.40\%\\
\quad$3H$&0.061&2.87\%&0.052&3.19\%\\
\quad$4H$&0.063&2.58\%&0.057&2.86\%\\
\hline
\end{tabular}
\end{center}
\end{table}

\begin{table}[b!]
\begin{center}
{
\caption{Confidence Interval (95\%)- Average performance of NRN with $1,\dots,~4$ hidden layers on $D_1$ through $D_{32}$.}
\label{allNets_ci}}
\begin{tabular}{lcccc}
\hline
\rule{0pt}{12pt}
&$Penetration$&$Width$\\
\hline
\\[-6pt]
\quad$1H$&(-0.001, 0.122)& (-0.0069, 0.116)\\
\quad$2H$&(0.007, 0.131)&(-0.0099, 0.114)\\
\quad$3H$&(-0.001, 0.122)&(-0.0099, 0.114)\\
\quad$4H$&(0.001, 0.125)&(-0.0049, 0.119)\\
\hline
\end{tabular}
\end{center}
\end{table}
\section{Scalability on Combined Data}
\label{further2}
This section provides result of performance of our model on combination of \emph{$D_1$} through \emph{$D_{5}$} to show its scalability on different welding techniques. We refer to this combined dataset as \emph{$D_{all}$}. Furthermore, we report the result of performance of our neural regression network (NRN) in contrast to normal equation regression (NER)~\cite{ChandelNFR1993},  multiple curvilinear regression (MCR)~\cite{karaoglusecgin2008,multiReg2007}, and support vector regression (SVR)~\cite{SVR2, SVR1}. The NER, using the pseudoinverse matrix of input feature vectors is~\cite{algorithmbook}:
\begin{align}
\begin{split}\label{normaleqn}
\Theta = (X^T X)^{-1} X^T Y
\end{split}
\end{align}
where \emph{X},\emph{Y}, and \emph{$\Theta$} are the IWP, DWP, and the weight matrix corresponding to all direct weld parameters, respectively. It is apparent that \emph{$\Theta$} is of \emph{$m \times \|DWP\|$} dimension with \emph{m} and \emph{$\|DWP\|$} being the size of training data and the cardinality of direct weld parameters. In addition, we use the Python scikit-learn package for modeling MCR and SVR~\cite{pythonpkg}.\newline
Tables~\ref{bestParamsOthersPenetration} and~\ref{bestParamsOthersWidth} show the best set of meta-parameters for NER, MCR, and SVR to calculate depth of penetration and width of weld beads, respectively. Hyphenated entries of these tables imply that the corresponding meta-parameters are not used by the given technique. In these tables, \emph{"Degree"} refers to the number of additional polynomial degrees to yield the best RMSEs of the DWP by a given model. It is selected from the range \emph{$\{0, 1,\dots, 6\}$}. Similarly, the \emph{kernel}, \emph{C}, and \emph{$\Gamma$} entries refer to the kernel function, the penalty parameter of the error term (equivalent to \emph{$\frac{1}{\alpha}$} in logistic regression), and the kernel coefficient that are associated with the SVR. The best kernel function is selected from possible choices of kernel for SVR in~\cite{pythonpkg}. They are \emph{rbf, ploy, sigmoid} and \emph{linear} kernels where \emph{rbf} and \emph{poly} stand for radial basis and polynomial functions, respectively. The kernel coefficient \emph{$\Gamma$} is associated with the first three of these functions.\newline
\begin{table}[b!]
\begin{center}
{\caption{Best meta-parameters for depth of penetration - Normal Equation Regression (NER), Multiple Curvilinear Regression (MCR), and Support Vector Regression (SVR) on \emph{$D_{all}$}.}
\label{bestParamsOthersPenetration}}
\begin{tabular}{lccccccc}
\hline
\rule{0pt}{12pt}
 &$\alpha$&$\lambda$&Degree&Kernel&C& $\Gamma$&\parbox[t]{1.0cm}{Iterations\\ $\in [1000,\dots\\, 12000]$}\\
\hline
\\[-6pt]
NER & - & - & 3 & - & - & - & ----- \\
MCR & 0.3 & 0.09 & 3 & - & - & - & 7000\\
SVR & - & - & 3 & poly & 0.009 & 0.001 & 5000\\
\hline
\end{tabular}
\end{center}
\end{table}
\begin{table}[b!]
\begin{center}
{\caption{Best meta-parameters for width of weld beads - NER, MCR, and SVR on \emph{$D_{all}$}.}
\label{bestParamsOthersWidth}}
\begin{tabular}{lccccccc}
\hline
\rule{0pt}{12pt}
 &$\alpha$&$\lambda$&Degree&Kernel&C& $\Gamma$&\parbox[t]{1.0cm}{Iterations\\ $\in [1000,\dots\\, 12000]$}\\
\hline
\\[-6pt]
NER & - & - & 2 & - & - & - & ----- \\
MCR & 0.03 & 0.09 & 2 & - & - & - & 4000\\
SVR & - & - & 0 & rbf & 300 & 0.03 & 1000\\
\hline
\end{tabular}
\end{center}
\end{table}
 \begin{table}[t!]
\begin{center}
{\caption{Best meta-parameters for depth of penetration (P) and width of weld beads (W) blocks -  NRN on \emph{$D_{all}$}.}
\label{metadrmpw}}
\begin{tabular}{lcccccc}
\hline
\rule{0pt}{12pt}
Datasets&\parbox[t]{1.0cm}{Neurons $\in [2,\dots\\, 100]$}&Degree&$\alpha$&$\gamma$&$\lambda$&\parbox[t]{1.0cm}{Iterations\\ $\in [1000,\dots\\, 12000]$}\\
\hline
\\[-6pt]
P & 9 & 0 & 0.001 & 3000 & 0.0 & 3000 \\
W & 7 & 0 & 0.001 & 2500 & 0.0 & 7000 \\
\hline
\end{tabular}
\end{center}
\end{table}
The best sets of meta-parameters for our neural regression model on \emph{$D_{all}$} is shown in Table~\ref{metadrmpw} . This table indicates that the polynomial degrees in our model remains zero, as its comparison to Tables~\ref{meta1} and~\ref{meta2} reveals. However, there is an increase in the number of neural units in the hidden layer of DWP blocks of our model, compared to their numbers, on average, in Tables~\ref{meta1} and~\ref{meta2} (i.e., maximum of 12 neurons in total with mean 9.50 and standard deviation 2.65). Moreover, the inverse proportionality of the values of learning rate \emph{$\alpha$} and reinforced learning coefficient \emph{$\gamma$} is, comparably, stronger in Table~\ref{metadrmpw} . These observations imply a higher variation in input feature vectors (i.e., IWP) to our model. However, such changes on meta-parameters of our model are expected as \emph{$D_{all}$} pertains to the combination of IWP of different welding techniques. Moreover, the regularization \emph{$\lambda$} indicates that our model is not susceptible to overfitting and bias on training data on \emph{$D_{all}$}. Furthermore, this observation holds true as we compare our neural regression model with \emph{MCR} and \emph{SVR}. This is evident in its \emph{$\lambda = 0.0$} as compared to \emph{C} and \emph{$\Gamma$} in Tables~\ref{bestParamsOthersPenetration} and~\ref{bestParamsOthersWidth}.\newline
\begin{table}[b!]
\begin{center}
{
\caption{RMSEs and PEs - NRN vs. NER, MCR, and SVR on $D_{all}$.}
\label{rmseALL2}}
\begin{tabular}{lcccccc}
\hline	
\rule{0pt}{12pt}
&Penetration (mm)&\multicolumn{3}{c}{Width}\\
\cline{1-5}
\rule{0pt}{12pt}
&$RMSE$&$PE$&$RMSE$&$PE$\\
\hline
\\[-6pt]
\quad NRN&0.153&3.50\%&0.146 & 2.87\%\\
\quad NER&0.343&6.88\%&0.895&11.20\%\\
\quad MCR&0.322&6.09\%&0.873&11.42\%\\
\quad SVR&1.586&31.81\%&0.400&8.02\%\\
\hline
\end{tabular}
\end{center}
\end{table}
Such adjustments on meta-parameters that are associated with the updates of weight matrices have positive influence on estimation of DWP as shown in Table~\ref{rmseALL2}. The RMSE entries of this table indicate that our model achieves 2.24 - 10.36 and 2.73 - 6.13 times improvements on estimation of the depth of penetration and width of weld beads, as compared to other models. Additionally, its \emph{z-score} for RMSE on width of weld beads shows a difference that is above one standard deviation to those of \emph{NER}, \emph{MCR}, and \emph{SVR} (their \emph{z-score} in the order of entries in Table~\ref{rmseALL2} are [-1.175, 0.860, 0.800, -0.485]). Furthermore, an analysis of the PE values in Table~\ref{rmseALL} indicates that these values are not within one standard deviation from each other (13.239 and 3.986 for depth of penetration and width of weld beads, respectively). Moreover, the non-overlapping confidence interval of our model with those of \emph{NER}, \emph{MCR}, and \emph{SVR} in Table~\ref{CICV_P} support the above observations. These analyses imply that the improvements exhibited by our model is statistically significant. \newline
Additionally, a comparison between the performance of our model on \emph{$D_{all}$} in contrast to Chandrasekhar and Vasudevan~\cite{chandrasekharvasudevan2010}, and Karaoglu and Secgin~\cite{karaoglusecgin2008} in Tables~\ref{t2008} through~\ref{t2010_2} indicates that the difference between these RMSE values are insignificant. More specifically, these RMSE values are within one standard deviation of one another (0.098, 0.003, and 0.020 for depth of penetration and 0.146, 0.041, and 0.007 for width of weld beads). Furthermore, this observation is supported by the overlapping confidence interval of our model and the intervals associated with these techniques in Table~\ref{CIDRMTHEM} . In other words, the estimates of DWP by our model based on combined data of different welding techniques is statistically as good as the results reported in~\cite{chandrasekharvasudevan2010} and~\cite{karaoglusecgin2008}. Furthermore, our model achieves better prediction errors on depth of penetration and width of weld beads while performing on \emph{$D_{all}$} than results reported in~\cite{andersonramaswamy1990}. A comparison between the PEs reported by these models indicates that the difference in their results are above one standard deviation (they are 11.37 and 1.98 for depth of penetration and width of weld beads, respectively). However, we cannot draw any conclusion on the difference of performance of our model on {$D_{all}$} and {$D_1$} with regrads to calculated RMSEs since Anderson et al.~\cite{andersonramaswamy1990} do not provide the RMSEs of their model on {$D_1$}.\newline
On the other hand, the quality of the calculated RMSE values of our model degrades when we compare its performance on \emph{$D_{all}$} in contrast to its results on \emph{$D_1$} through \emph{$D_{5}$}. The \emph{z-score} of its calculated RMSEs on depth of penetration are [0.432, -0.632, 0.557, -0.569, -0.799, -0.841, 1.852]. However, the standard deviation of these \emph{z-scores} equals 1.00 which is still within one standard deviation from each other. Same observation holds true in case of calculated RMSEs for width of weld beads, with their respective \emph{z-scores} equal [0.243, -0.570, 0.116, -0.570, -0.671, -0.646, 2.099]. It is worth noting that the overlapping confidence interval of \emph{$D_{all}$} with entries in Table~\ref{CIDRM} i.e., the confidence inetrval of our model on individual datasets \emph{$D_1$} through \emph{$D_5$} at $95\%$ level suggests that the change in performance of our minimalist neural regression model is insignificant.
\begin{table}[t!]
\begin{center}
{
\caption{Confidence Interval (95\% - $D_{all}$) - NRN vs. NER, MCR, and SVR.}
\label{CICV_P}}
\begin{tabular}{lcccc}
\hline
\rule{0pt}{12pt}
&$Penetration$&$Width$\\
\hline
\\[-6pt]
\quad NRN&(0.128, 0.252)& (0.130, 0.274)\\
\quad NER&(0.281, 0.405)&(0.833, 0.957)\\
\quad MCR&(0.280, 0.384)&(0.811, 0.935)\\
\quad SVR&(1.524, 1.648)&(0.338, 0.462)\\
\hline
\end{tabular}
\end{center}
\end{table}

\begin{table}[b!]
\begin{center}
{
\caption{Confidence Interval (95\%) - NRN vs.~\cite{chandrasekharvasudevan2010} and~\cite{karaoglusecgin2008}.}
\label{CIDRMTHEM}}
\begin{tabular}{lcccc}
\hline
\rule{0pt}{12pt}
&$Penetration$&$Width$\\
\hline
\\[-6pt]
\quad $D_{all}$& (0.128, 0.252)& (0.130, 0.274)\\
\quad $D_2$ & (0.230, 0.354)&(0.246, 0.415)\\
\quad $D_{31}$ & (0.086, 0.211)&(0.143, 0.267)\\
\quad $D_{32}$ & (0.062, 0.186)&(0.094, 0.218)\\
\hline
\end{tabular}
\end{center}
\end{table}

\begin{table}[t!]
\begin{center}
{
\caption{Confidence Interval (95\%) - NRN.}
\label{CIDRM}}
\begin{tabular}{lcccc}
\hline
\rule{0pt}{12pt}
&$Penetration$&$Width$\\
\hline
\\[-6pt]
\quad $D_{all}$& (0.091, 0.215)& (0.084, 0.208)\\
\quad $D_1$&(0.019, 0.143)&(0.031, 0.155)\\
\quad $D_2$&(-0.031, 0.102)&(-0.040, 0.084)\\
\quad $D_{31}$&(0.051, 0.175)&(-0.008, 0.120)\\
\quad $D_{32}$&(-0.019, 0.110)&(-0.006, 0.118)\\
\quad $D_{4}$&(-0.017, 0.108)&(-0.004, 0.113)\\
\quad $D_{5}$&(-0.015, 0.098)&(-0.004, 0.114)\\
\hline
\end{tabular}
\end{center}
\end{table}

\section{Conclusion}
\label{conclude}
We present a minimalist neural regression architecture as an aggregate of number of independent neural regression blocks where number of these blocks is proportional to the cardinality of the direct weld parameters (DWP). We show that reinforcing the calculated gradients during the backpropagation along with the weighted estimates using the learned mean of discrepancies results in a significant performance improvement in comparison with state-of-the-art techniques in the literature. In addition, we show that our model retains the quality of its estimates while presented with datasets of different welding techniques. This is an interesting and nontrivial observation since formulation of a model with scalability on estimating DWP of different welding techniques is highly desirable. Moreover, we show that the increase in the number of hidden layers of our model does not yield a significant improvement on its performance. In other words, its minimalist design does not jeopardize the quality of its estimation.\newline
Although the adapted sets of meta-parameters of the DWP blocks of our model indicates its ability to realize the high variation of the input feature vectors (IWP), a comparison between its performance on combined data of different welding techniques versus its results on separate datasets reveals a decay in its estimations accuracy (in the form of calculated RMSE and PE values). Therefore, future of this research pertains to further analysis of this degradation of the performance of our model on combined data to realize the potential solutions for resolving this behaviour.\newline
In addition, we search for more publicly available welding data with whose estimation approaches not included in our comparative study to broaden the domain of welding techniques that are investigated in this study to evaluate the effect of the size of the training data on the quality as well as scalability of our model.\newline
Moreover, most welding tasks are multi-layer welding processes. This results in every preceding layer of welded material to affect the geometry of the beads of the succeeding layer due to heat, their surface layout, and so on. Therefore, it is crucial to analyze the performance of our model on such multi-layer welding to determine the potential that it can offer to the solution concept of these operations.\newline
Estimation of the direct weld parameters represents a challenging problem in industrial domain. However, it is insufficient to draw a decisive conclusion on effectiveness of our proposed minimalist neural regression model based on reinforced learning coefficient and weighted estimates. As a result, it is crucial to investigate the applicability of our model in other domain of regression estimation, thereby realizing its utility.

%
%


\bibliographystyle{IEEEtran}
\bibliography{refs}

\end{document}